%% file: 0.POVC.tex
\documentclass[sigconf]{acmart}

\AtBeginDocument{%
  \providecommand\BibTeX{{%
    \normalfont B\kern-0.5em{\scshape i\kern-0.25em b}\kern-0.8em\TeX}}}

\copyrightyear{2020}
\acmYear{2020}
\setcopyright{acmcopyright}\acmConference[MM '20]{Proceedings of the 28th ACM International Conference on Multimedia}{October 12--16, 2020}{Seattle, WA, USA}
\acmBooktitle{Proceedings of the 28th ACM International Conference on Multimedia (MM '20), October 12--16, 2020, Seattle, WA, USA}
\acmPrice{15.00}
\acmDOI{10.1145/3394171.3413880}
\acmISBN{978-1-4503-7988-5/20/10}
\usepackage{multirow}
\usepackage{footmisc}

\begin{document}

%%
%% The "title" command has an optional parameter,
%% allowing the author to define a "short title" to be used in page headers.
\title{Poet: Product-oriented Video Captioner for E-commerce}

%
% The "author" command and its associated commands are used to define
% the authors and their affiliations.
% Of note is the shared affiliation of the first two authors, and the
% "authornote" and "authornotemark" commands
% used to denote shared contribution to the research.
\author[S. Zhang*, Z. Tan*, J. Yu, Z. Zhao, K. Kuang, J. Zhou, H. Yang, F. Wu]{
    Shengyu Zhang$^{1*}$, Ziqi Tan$^{1*}$, Jin Yu$^{2}$, Zhou Zhao$^{1\dagger}$, Kun Kuang$^{1\dagger}$, Jie Liu$^{2}$, Jingren Zhou$^{2}$, Hongxia Yang$^{2\dagger}$, Fei Wu$^{1\dagger}$
}
\affiliation{
    $^1$ College of Computer Science and Technology, Zhejiang University
}
\affiliation{
    $^2$ Alibaba Group
}
\email{
  {sy_zhang, tanziqi, zhaozhou, kunkuang, wufei}@zju.edu.cn
}
\email{
  {kola.yu, sanshuai.lj, jingren.zhou, yang.yhx}@alibaba-inc.com
}
\renewcommand{\authors}{Shengyu Zhang, Ziqi Tan, Jin Yu, Zhou Zhao, Kun Kuang, Jie Liu, Jingren Zhou, Hongxia Yang, Fei Wu}

%%
%% By default, the full list of authors will be used in the page
%% headers. Often, this list is too long, and will overlap
%% other information printed in the page headers. This command allows
%% the author to define a more concise list
%% of authors' names for this purpose.
%\renewcommand{\shortauthors}{S. Zhang*, Z. Tan*, J. Yu, Z. Zhao, K. Kuang, T. Jiang, J. Zhou, H. Yang, F. Wu}
\newcommand{\model}{ model }
\newcommand{\etal}{\textit{et al}.}
\newcommand{\ie}{\textit{i}.\textit{e}.}
\newcommand{\eg}{\textit{e}.\textit{g}.}
\newcommand{\Our}{\textit{Poet}   }
\newcommand{\vpara}[1]{\vspace{0.05in}\noindent\textbf{#1 }}

%%
%% The abstract is a short summary of the work to be presented in the
%% article.
\begin{abstract}

In e-commerce, a growing number of user-generated videos are used for product promotion. How to generate video descriptions that narrate the user-preferred product characteristics depicted in the video is vital for successful promoting. Traditional video captioning methods, which focus on routinely describing what exists and happens in a video, are not amenable for \textit{product-oriented} video captioning. To address this problem, we propose a product-oriented video captioner framework, abbreviated as \textit{Poet}. \textit{Poet} firstly represents the videos as product-oriented spatial-temporal graphs. Then, based on the aspects of the video-associated product, we perform knowledge-enhanced spatial-temporal inference on those graphs for capturing the dynamic change of fine-grained product-part characteristics. The knowledge leveraging module in \textit{Poet} differs from the traditional design by performing knowledge filtering and dynamic memory modeling. We show that \textit{Poet} achieves consistent performance improvement over previous methods concerning generation quality, product aspects capturing, and lexical diversity. Experiments are performed on two product-oriented video captioning datasets, buyer-generated fashion video dataset (BFVD) and fan-generated fashion video dataset (FFVD), collected from Mobile Taobao. We will release the desensitized datasets to promote further investigations on both video captioning and general video analysis problems.

%In this paper, we propose to investigate the real-world problem, \textit{video tone}, to automatically narrate the user-preferred product characteristics, appearing in e-commerce user-generated videos, in natural language. Different from traditional video-to-language problems (\eg, video captioning) that mainly recognize the main objects and general activities, \textit{video tone} naturally requires a fine-grained product-oriented video analysis as well as the ability to incorporate the external knowledge, \ie, video-associated product aspects. To this end, we firstly propose to represent the videos as product-oriented spatial-temporal graphs. We then propose a novel framework, named \textit{Keeper}, to progressively perform knowledge-enhanced spatial-temporal inference. The knowledge leveraging module in \textit{Keeper} differs from the traditional design by performing knowledge filtering and dynamic memory modeling. We show that \textit{Keeper} achieves consistent performance improvement over previous methods concerning generation quality, product aspects capturing, and lexical diversity. Experiments are performed on two \textit{video tone} datasets, \textit{BFVD} and \textit{FFVD}, collected from the real-world e-commerce platform, Taobao. We will release the desensitized version to promote further investigations on both \textit{video tone} and general video analysis problems.

\end{abstract}

%%
%% The code below is generated by the tool at http://dl.acm.org/ccs.cfm.
%% Please copy and paste the code instead of the example below.
%%



%\begin{CCSXML}
%<ccs2012>
% <concept>
%  <concept_id>10010520.10010553.10010562</concept_id>
%  <concept_desc>Computer systems organization~Embedded systems</concept_desc>
%  <concept_significance>500</concept_significance>
% </concept>
% <concept>
%  <concept_id>10010520.10010575.10010755</concept_id>
%  <concept_desc>Computer systems organization~Redundancy</concept_desc>
%  <concept_significance>300</concept_significance>
% </concept>
% <concept>
%  <concept_id>10010520.10010553.10010554</concept_id>
%  <concept_desc>Computer systems organization~Robotics</concept_desc>
%  <concept_significance>100</concept_significance>
% </concept>
% <concept>
%  <concept_id>10003033.10003083.10003095</concept_id>
%  <concept_desc>Networks~Network reliability</concept_desc>
%  <concept_significance>100</concept_significance>
% </concept>
%</ccs2012>
%\end{CCSXML}
%
%\ccsdesc[500]{Computer systems organization~Embedded systems}
%\ccsdesc[300]{Computer systems organization~Redundancy}
%\ccsdesc{Computer systems organization~Robotics}
%\ccsdesc[100]{Networks~Network reliability}

%%
%% Keywords. The author(s) should pick words that accurately describe
%% the work being presented. Separate the keywords with commas.


%% A "teaser" image appears between the author and affiliation
%% information and the body of the document, and typically spans the
%% page.
%\begin{teaserfigure}
%  \includegraphics[width=\textwidth]{sampleteaser}
%  \caption{Seattle Mariners at Spring Training, 2010.}
%  \Description{Enjoying the baseball game from the third-base
%  seats. Ichiro Suzuki preparing to bat.}
%  \label{fig:teaser}
%\end{teaserfigure}

%%
%% This command processes the author and affiliation and title
%% information and builds the first part of the formatted document.
\maketitle

\renewcommand{\thefootnote}{\fnsymbol{footnote}}
\footnotetext[1]{These authors contributed equally to this work.}
\footnotetext[2]{Corresponding Authors.}
\footnotetext{Work was performed when S. Zhang and Z. Tan were interns at Alibaba Group.}
\renewcommand{\thefootnote}{\arabic{footnote}}
\fancyhead{}

\input{1.Introduction}

\input{2.Related_Works}

%\include{3.DataCollection}

\input{4.Method}

\input{5.Experiments}

\input{6.Conclusion}

\bibliographystyle{ACM-Reference-Format}
\bibliography{9.citations}

\appendix

\end{document}

%% file: 1.Introduction.tex
\section{Introduction}

Nowadays, a growing number of short videos are generated and uploaded to Taobao. Among these videos, user-generated videos are massive in volume and share unique product experiences, such as the individual preference for the product usage scenario or usage strategy. When recommending these videos to customers for product promotion, accompanying a description that narrates the uploader-preferred highlights depicted in the product video is essential for successful promotion, \ie, attracting potential customers with similar interests or preferences to buy the same product. Different from traditional video-to-text generation problems which mainly concern what exists or happens in the video, this problem cares about what the video uploader wants to highlight. We name this particular problem as \textit{product-oriented} video captioning.

\begin{figure}[!t] \begin{center}
    \includegraphics[width=0.9\columnwidth]{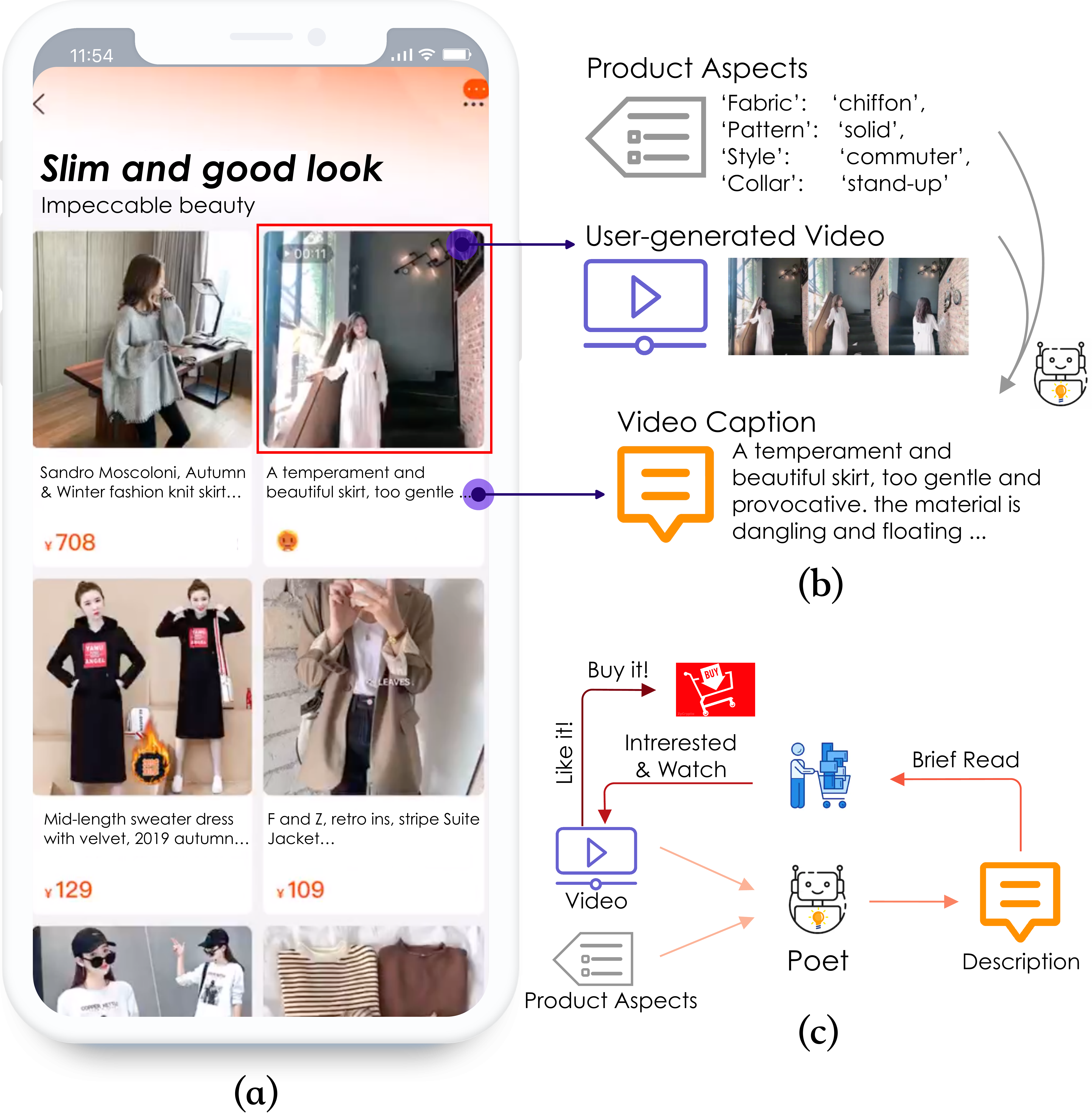}
    \caption{
    An illustration of \textit{product-oriented} video captioning. (a) A real-world product video recommendation scenario. (b) A showcase of the product aspects, user-generated video, and a video caption. (c) An illustration of how \textit{Poet} contribute to product promotion.
    	}
\vspace{-0.4cm}
\label{fig:application}
\end{center} \end{figure}
%--------------------------------fig end--------------------

\begin{sloppypar}
	Product-oriented video captioning naturally requires a fine-grained analysis of prominent product characteristics depicted in the video. However, without some general understanding of the product, it can be hard even for a human to grasp what the uploader mainly concerns based on the isolated video. To this end, we view leveraging product-related knowledge as a fundamental ability for product-oriented video captioning. Concretely, we take the structured product aspects from the associated product as prior knowledge since they are easy to acquire (most user-generated videos in e-commerce have on-sell product associations.) and concise in meaning. The structured product aspects arranged by product sellers contain general and basic information necessary for fine-grained video understanding. Figure \ref{fig:application} reveals the task definition, the application scenario in Mobile Taobao, and how automatic tools contribute to product promotion \footnote{\label{foot1}For the original Chinese descriptions as well the desensitized datasets, please refer to \url{https://github.com/shengyuzhang/Poet}}.
\end{sloppypar}

\iffalse

这两个原因其实是同一个，不能加入product的信息，就做不到product oriented
通过video caption和video tone两个任务的本质区别，提出video tone的挑战，比如： 1) how to incorporate the product information for product oriented video tone; 2) how to model spatial interactions between region-region and region-background within frames 等
Video tone是否可以认为是producte oriented video caption??

\fi

%\textit{Video tone} can be more generally formulated as a video-language translation problem, typically video captioning. 

Recent advances in deep neural networks \cite{Wang_Wang_Huang_Wang_Tan_2018,Liu_Ren_Yuan_2018,Tang_Wang_Wang_Xu_2017}, especially the RNNs, have convincingly demonstrated high capability in handling the general video captioning problem. Most of them \cite{Venugopalan_Rohrbach_Donahue_Mooney_Darrell_Saenko_2015,Yao_Torabi_Cho_Ballas_Pal_Larochelle_Courville_2015,Pan_Xu_Yang_Wu_Zhuang_2016} incorporate an RNN based encoder-decoder structure with/without attention mechanisms to perform sequential frames encoding and sequential words decoding. However, \textit{product-oriented} video captioning poses some unique challenges. A first challenge comes from mining the product characteristics from the user-generated product videos and capturing the dynamic interactions of these characteristics within and across frames, as well as exploring the interactions between the product and the background scene. We name this fundamental requirement as product-oriented video modeling, which is essential for discovering what the uploader highlights. Another challenge relates to the joint modeling of discrete product aspects and the video. The cross-modal nature of these inputs makes the modeling even more challenging.

%these architectures are not amenable to our task since these methods are not \textit{product-oriented}.
%
%Among the major challenges of \textit{product-oriented} video captioning are the capacity to discover

%On the one hand, these methods are incapable of capturing the fine-grained product characteristics in videos or the interactions between these details since they mainly incorporate the RNNs to model the sequential dependency between video frames. On the other hand, they are incapable of modeling the external background knowledge, \ie, product aspects, for product-oriented video analysis.

%
%
%for the following three reasons: 1) These methods solely model the video information in the frame-level. These coarse-grained representations may not well deliver the preferred highlights of product characteristics. 2) Though RNN based encoders may well model the sequential dependencies between video frames, they are incapable of capturing the fine-grained product characteristics, which can have special dependencies beyond the markovian dependence. 3) These methods are designed for video analysis only and may fail to exploit external knowledge or to bridge the gap between different modalities. 

To address these challenges, we propose an approach, named \textbf{P}roduct-\textbf{o}riented Vid\textbf{e}o Cap\textbf{t}ioner and abbreviated as \textbf{\textit{Poet}}. For product-oriented video modeling, \textit{Poet} first represents the product parts as graph nodes and build a spatial-temporal video graph. Then, \textit{Poet} encapsulates the proposed spatial-temporal inference module to perform long-range temporal modeling of product parts across frames and spatial modeling of product parts within the same frame. We also mix a spatial frame node into the video graph to explore the interactions between the foreground (product) and the background (scene). For the second challenge, we propose the knowledge leveraging module, which comprises the knowledge filtering process and knowledge aggregation process.

% By representing the video as a product-oriented spatial-temporal video graph, \textit{Poet} pe rforms long-range temporal modeling of product parts across frames and spatial modeling of product parts within the same frame. \textit{Poet} encapsulates the spatial-temporal inference module for information propagation on the graph.

%This module is based on the flexible graph neural network, and we differ from the vanilla version by separately modeling the root node and neighbors nodes during information aggregation and employing a position-wise score function for importance re-weighting. To obtain a better understanding of the product-oriented videos with product aspects, we propose the knowledge leveraging module, which comprises the knowledge filtering process and knowledge aggregation process based on the memory network. We differ from the vanilla end-to-end memory network \cite{Sukhbaatar_Szlam_Weston_Fergus_2015} by performing knowledge filtering and dynamic memory writing.

To accommodate the \textit{product-oriented} video captioning research, we build two large-scale datasets, named buyer-generated fashion video dataset (BFVD) and fan-generated fashion video dataset (FFVD), from Mobile Taobao. There are 43,166 and 32,763 <video, aspects, description> triplets in BFVD and FFVD, respectively. On the language side, descriptions in these two datasets have an extensive vocabulary and considerable product details. Compared with captions in existing video datasets that mostly describe the main objects and the overall activities, descriptions in BFVD and FFVD reflect the characteristics of product-parts, the overall appearance of the product, and interactions between products and backgrounds. Such new features present unique challenges for \textit{product-oriented} video captioning and also for general video analysis.

To summarize, this paper makes the following key contributions:

\begin{sloppypar}
	\begin{itemize}
	\item We propose to investigate a real-world video-to-text generation problem, \textit{product-oriented} video captioning, to automatically narrate the user-preferred product characteristics inside user-generated videos.
	\item We propose a novel Poet framework for \textit{product-oriented} video captioning via simultaneously capturing the user-preferred product characteristics and modeling the dynamic interactions between them with a product-oriented spatial-temporal graph.
	\item We introduce a novel knowledge leveraging module to incorporate the product aspects for product video analysis by performing knowledge filtering, dynamic memory writing, and knowledge attending. \textit{Poet} yields consistent quantitative and qualitative improvements on two \textit{product-oriented} video captioning datasets \footref{foot1} that are collected in Mobile Taobao.
\end{itemize}

\end{sloppypar}

%% file: 2.Related_Works.tex
\section{Related Works}

\subsection{Video to Text Generation}

Most deep learning based video-to-text generation methods \cite{Venugopalan_Xu_Donahue_Rohrbach_Mooney_Saenko_2015,Venugopalan_Rohrbach_Donahue_Mooney_Darrell_Saenko_2015,Yao_Torabi_Cho_Ballas_Pal_Larochelle_Courville_2015,Pan_Xu_Yang_Wu_Zhuang_2016,Wang_Ma_Zhang_Liu_2018,Ma_Cui_Dai_Wei_Sun_2019,Shi_Cai_Joty_Gu_2019,Zhu_Jiang_2019,Barati_Chen_2019,Xu_Yao_Zhang_Mei_2017,Yang_Xu_Wang_Wang_Han_2017,zhou2020progress} focus on sequence-to-sequence modeling and employ RNN based encoder-decoder structures. Typically, S2VT \cite{Venugopalan_Rohrbach_Donahue_Mooney_Darrell_Saenko_2015} firstly formulates video-to-text as a sequence to sequence process. SALSTM \cite{Yao_Torabi_Cho_Ballas_Pal_Larochelle_Courville_2015} improves the decoding process using the effective soft-attention mechanism. HRNE \cite{Pan_Xu_Yang_Wu_Zhuang_2016} builds a hierarchical RNN design for representing videos. RecNet \cite{Wang_Ma_Zhang_Liu_2018} proposes to re-produce the input frames features after decoding. %Though they have achieved superior performance improvement over previous methods by exploiting frame-level features and sequential dependencies, they have limitations in capturing fine-grained spatial cues inside each frame.

% Unlike general video to Text generation problems, which concerns recognizing main objects and the overall object activities,

% only model the temporal structure of the regions across frames and

More recently, there are works exploiting object-level features in representing the videos \cite{Zhang_Peng_2019,Yang_Han_Wang_2017,Hu_Chen_Zha_Wu_2019,Wang_Xu_Han_2018} for video description generation. They mainly propose to detect salient objects and employ RNNs to model the temporal structure between them. However, they are not directly applicable to \textit{product-oriented} video captioning for the following two reasons: 1) \textit{product-oriented} video captioning requires even more fine-grained video analysis, \ie, product-part characteristic recognition. 2) These methods neglect the spatial interactions between region-region and region-background within frames. %Background here denotes the undetected visual regions, which are essential for figuring out the highlights of videos. For example, customers wearing a jacket in the rain without an umbrella may want to show the waterproof performance. To this end, 
\textit{Poet} represents both the detected product-parts and the whole frames as spatial-temporal graphs and employs the graph neural network to model the interactions between product-parts and product-background.

\subsection{Knowledge enhanced Video Analysis}

% , which helps obtain a more comprehensive understanding about the main object or the event.

Incorporating external in-domain knowledge is a promising research direction \cite{pan2019multiple} for video analysis. There are mainly two kinds of external knowledge, \ie, knowledge graph and topically related documents (\eg, Wikipedia). Knowledge graph based methods \cite{Gao_Zhang_Xu_2019,Gao_Zhang_Xu_2018} typically retrieve the knowledge graph from off-the-shelf knowledge bases such as ConceptNet 5.5 \cite{Speer_Chin_Havasi_2017} and employ the graph convolution network \cite{Kipf_Welling_2017} to perform knowledge reasoning. These methods are not suitable for our task since there are no well-defined relationships among product aspects. For document-based approaches, Venugopalan \etal \cite{Venugopalan_Hendricks_Mooney_Saenko_2016} uses the Wikipedia corpus to pre-train a language model (LM) and proposes the late/deep fusion strategies to enhance the decoding RNN with the LM. Whitehead \etal \cite{Whitehead_Ji_Bansal_Chang_Voss_2018} first retrieves the relevant document and then use the pointer mechanism to directly borrow entities in the decoding stage. Different from these works, \textit{Poet} performs knowledge leveraging in the product-oriented spatial-temporal inference stage. % to obtain a better understanding and encoding of the visual cues rather than perform direct aspects-attending in the decoding stage. %This is especially designed for \textit{product-oriented} video captioning, which focuses more on recognizing preferred product characteristics appearing in the user-generated videos.

%  pre-training language dataset for the decoder RNN, and fuse the attention weights on entities (extracted from related documents) and the predicted vocabulary propabilities in the decoding stage.

%% file: 4.Method.tex
%--------------------------------fig---------------------
\begin{figure*}[t] \begin{center}
    \includegraphics[width=0.9\textwidth]{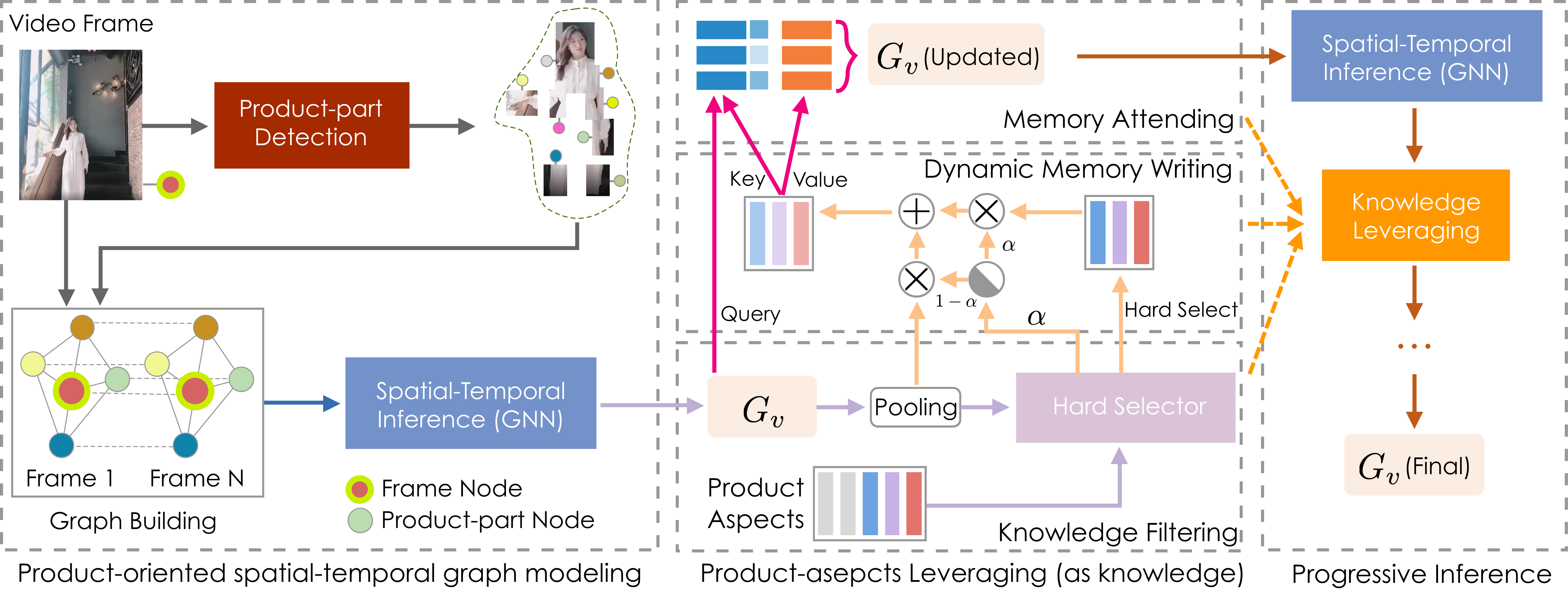}
    \caption{
    Schematic of the proposed \textit{Poet} encoder. We represent the video as a product-oriented spatial-temporal graph with product parts and frames as nodes. The \textit{Poet} encoder progressively performs spatial-temporal inference and knowledge leveraging to obtain a better understanding of the video with the product aspects. The knowledge leveraging module incorporates three sub-processes, \ie, knowledge filtering, dynamic memory writing, and memory attending.
	}
\vspace{-0.3cm}
\label{fig:schema}
\end{center} \end{figure*}
%--------------------------------fig end---------------------
%To clarify, GC stands for Graph Convolutions and GLA for global-local aggregation.

\section{Methods}

%%--------------------------------fig-------------------------
%\begin{figure}[!t] \begin{center}
%    \includegraphics[width=\columnwidth]{./figures/GLA.pdf}
%    \caption{
%    A detailed illustration of the proposed global-local aggregation module for aligning and combining heterogeneous graphs. We firstly encourage context information relevant to the global representation of the query graph (global gated access) and further attend to the context graph for each query node (local attention).
%	}
%\vspace{-0.4cm}
%\label{fig:GLA}
%\end{center} \end{figure}
%%--------------------------------fig end--------------------

\subsection{Overview}

After data preprocessing (details in Section \ref{Data Collection}), we represent each product-oriented video as frame-level features $\{\mathbf{f}_{i} \in \mathbb{R}^{D_f} \}_{i = 1,\dots,N_f}$ where $\mathbf{f}_{i}$ is the $D_v$ length feature vector for the $i$th frame $f_i$, and product-part features $\{ \mathbf{p}_{i,j} \in \mathbb{R}^{D_p} \}_{j=1,\dots,N_p}$ where $\mathbf{p}_{i,j} $ is the $D_p$ length feature vector for the $j$th product-part in the $i$th frame. The video-associated product aspects are $ \{ a_k \}_{k = 1,\dots,N_a} $ and we use the embedding layer to learn an aspect embedding $\mathbf{a}_k \in \mathbf{R}^{D_a}$ of dimension $D_a$ for the $k$th aspect $a_k$. We aim to generate a video description $\{w_m\}_{m=1,\dots,N_w}$ that narrates the preferred product characteristics of e-commerce buyers/fans.

%After data preprocessing (details in Section \ref{Data Collection}), we represent each product-oriented video as frame-level features $\{\mathbf{f}_{i} \in \mathbb{R}^{D_f} \}_{i = 1,\dots,N_f}$ and product-part features $\{ \mathbf{p}_{i,j} \in \mathbb{R}^{D_p} \}_{j=1,\dots,N_p}$. $\mathbf{f}_{i}$ is the $D_v$ length feature vector for the $i$th frame $f_i$ and $\mathbf{p}_{i,j} $ is the $D_p$ length feature vector for the $j$th product-part in the $i$th frame. The video-associated product aspects are $ \{ a_k \}_{k = 1,\dots,N_a} $ and we use the embedding layer to learn an aspect embedding $\mathbf{a}_k \in \mathbf{R}^{D_a}$ of dimension $D_a$ for the $k$th aspect $a_k$. We aim to generate a video description $\{w_m\}_{m=1,\dots,N_w}$ that narrates the preferred product characteristics of e-commerce buyers/fans.

We firstly build a product-oriented spatial-temporal video graph (See Figure \ref{fig:schema}), which contains both frame nodes and product-part nodes. With the graph representation, the encoder of \textit{Poet} mainly incorporates two sub-modules, \ie, the spatial-temporal inference module for graph modeling, and the knowledge leveraging module for product aspects modeling. These sub-modules can be easily stacked to obtain a progressive knowledge retrieval and knowledge enhanced visual inference process. In the next several subsections, we will formally introduce the building blocks comprising \textit{Poet}, including the graph building process, the spatial-temporal inference module, the knowledge leveraging module, and the attentional RNN-based decoder in detail.

\subsection{Product-oriented Video Graph Modeling}

\subsubsection{Graph Building} To better capture the highlights (\ie, preferred product characteristics) inside the product videos, we propose to represent the videos as spatial-temporal graphs.

\vpara{Nodes} Different from previous works \cite{Zhang_Peng_2019,Yang_Han_Wang_2017,Hu_Chen_Zha_Wu_2019,Wang_Xu_Han_2018} that represent the objects as graph nodes, we represent product parts as nodes to capture the dynamic change of these fine-grained details along the timeline. Product-part features are extracted by a pre-trained CNN-based detector (details in \ref{Data Collection}), and thus these features naturally contain spatial cues. However, since we do not model the product parts along the timeline using RNNs, there is no concept of frame order in the modeling process. To this end, we add an order-aware embedding $\mathbf{o}_i \in \mathbb{R}^{D_p}$ to each product-part feature, which is similar to the position embedding strategy employed in sequence learning \cite{gehring2017convolutional,lu2019vilbert}. $\mathbf{o}_i$ stands for the embedding for the frame order $i$. Sharing similar spirit, we further obtain the type-aware product-part representation by adding the part-type embedding $\mathbf{t}_j \in \mathbb{R}^{D_p}$, which stands for the $j$th part for a particular product, such as waistline and hem. Then, the enhanced product-part feature $\mathbf{p}^e_{i,j}$ can be obtained by:
\begin{align}
	\mathbf{p}^e_{i,j} = \mathbf{p}_{i,j} + \mathbf{o}_i + \mathbf{t}_j,
\end{align}
Besides the product-part nodes, we further incorporate the frame node into each frame graph to capture the product as a whole and exploit the correlations between the products and the backgrounds. Similar to the product-part feature, we add the order-based embedding $\mathbf{o}_i$ and a special type embedding $\mathbf{t}_{[\textrm{frame}]}$ for obtaining the frame-order concept and the frame-type concept, respectively.
\begin{align}
	\mathbf{f}^e_{i} = \mathbf{f}_{i} + \mathbf{o}_i + \mathbf{t}_{[\textrm{frame}]}.
\end{align}
We then project the product-part features and the frame features into a common space by employing two linear transformations, \ie, $\mathbf{W}_p \in \mathbb{R}^{D_{pf} \times D_{p}}$ and $\mathbf{W}_f \in \mathbb{R}^{D_{pf} \times D_{f}}$.
\begin{align}
	\mathbf{v}^p_{i,j} = \mathbf{W}_p \mathbf{p}^e_{i,j} + \mathbf{b}_p, \ \ \mathbf{v}_{i, [\textrm{frame}]} = \mathbf{W}_f \mathbf{f}^e_{i} + \mathbf{b}_f.
\end{align}
$\mathbf{b}_p, \mathbf{b}_f \in \mathbb{R}^{D_{pf}}$ are the bias terms. Therefore, each frame graph contains $N_p$ product-part nodes $\{ v^p_{i,j} \}_{j=1,\dots,N_p}$ with representations $\{ \mathbf{v}^p_{i,j} \}_{i=1,\dots,N_p}$ and one frame node $v_{i,[\textrm{frame}]}$ with representation $\mathbf{v}_{i,[\textrm{frame}]}$. There are totally $N_f$ frame sub-graphs in the whole graph.

\vpara{Edges} To capture the correlations among product-parts within the same frame as well as the interactions between global frame context (\eg, background, the product as a whole) and local part details, we propose a baseline method by fully connecting the product-part nodes and frame node within each frame graph. To obtain a comprehensive understanding and the dynamic change of product-parts across different viewpoints, we fully connect the nodes of the same type (including the frame type) from all frames. The edge weights are obtained using a fully-connected layer:
\begin{align}
	\mathbf{e}_{\iota,\kappa} = \mathbf{W}_e [\mathbf{v}_{\iota}, \mathbf{v}_{\kappa}] + \mathbf{b}_e,
\end{align}
where $\mathbf{e}_{\iota,\kappa} \in \mathbb{R}$ is the weight of edge between node $v_{\iota}$ and $v_{\kappa}$. We use a linear transformation $\mathbf{W}_e \in \mathbb{R}^{1 \times 2*D_{pf}}$ with a bias term $\mathbf{b}_e \in \mathbb{R}$ to estimate the correlation of two nodes. $[,]$ denotes the concatenation operation. For convenience, we use $G_v$ to denote the initial video graph and intermediate video graphs since only nodes feature representations are updated.

%\begin{table}[t]
%%\begin{strip}
%\centering
%\caption{
%    %
%    %
%    %
%    Notations.
%}
%\setlength\doublerulesep{0.5pt}
%%\footnotesize  
%\small
%\begin{tabular}{c|l}
%%\hline
%%\toprule
%%& \multicolumn{4}{c}{Referring Expression Generation}  \\
%%\cline{1-9}
%% <<<
%\hline
%\hline
%\multicolumn{1}{c|}{ \textbf{Notation} }        & \multicolumn{1}{l}{ \textbf{Description} }          \\
%%\midrule
%\hline \hline
%$\mathbf{f}, \mathbf{p}, \mathbf{a}$     & feature vector of frames/product parts/product aspects \\
%$\mathbf{o}, \mathbf{t}$     & position/node-type embeddings \\
%$\mathbf{v}^f, \mathbf{v}^p$ & feature vector of frame nodes / product-part nodes \\
%$\mathbf{e_{\iota,\kappa}}$ & the edge weight between the $\iota$th node and $\kappa$th node \\
%$\mathbf{W}, \mathbf{b}$ & the matrix / bias term of linear transformation \\
%$N$ & the number of elements in a sequence/set \\
%$D$ & the dimension of feature vectors \\
%$\mathbf{h}_t$ & feature vector of RNN hidden state at step $t$ \\ 
%$\mathbf{\alpha}, \mathbf{\omega}, \mathbf{\varrho}$ & attention weights \\
%\hline
%% >>>
%\end{tabular}
%%\medskip
%\vspace{-0.4cm}
%\label{tab:notations}
%\end{table}

\subsubsection{Spatial-temporal Inference} Although previous works \cite{Yang_Han_Wang_2017,Hu_Chen_Zha_Wu_2019,Wang_Xu_Han_2018} have proposed to capture the fine-grained region-of-interests such as objects and \cite{Zhang_Peng_2019} proposes to represent these fine-grained cues as graphs, they all use RNN-based modeling which can be inefficient for its internally recurrent nature and can be less effective in modeling interactions of regions within a frame since these regions have no natural temporal dependencies. To this end, we employ the flexible graph neural networks for spatial-temporal inference. Existing works performing video graph modeling for video relation detection \cite{Qian_Zhuang_Li_Xiao_Pu_Xiao_2019}, temporal action localization \cite{Zeng_Huang_Tan_Rong_Zhao_Huang_Gan_2019}, and video action classification \cite{Wang_Gupta_2018} often leverage the off-the-shelf Graph Convolutional Networks \cite{kipf2016semi} for information propagation. We propose a new modeling schema by separately modeling the root node and neighbor nodes when aggregation. For neighbor nodes information aggregation, we use an element-wise max function for its effectiveness in the experiment:
%\begin{align}
%	\mathbf{\bar v}_{\iota}^n = \max_{1\leq k\leq N_{i}} \{\mathbf{e}_{\iota,\kappa} * \mathbf{v}_\kappa, v_\kappa \in \mathcal{N}(v_\iota) \}, \mathrm{where} \ \mathbf{v}_\kappa = \{ \mathbf{v}_{\kappa, k} \}_{k=1,\dots,N_{i}}
%\end{align}
\begin{align}
	\mathbf{\bar v}_{\iota, \varsigma}^n = \max_{\kappa} \{\mathbf{e}_{\iota,\kappa} * \mathbf{v}_{\kappa, \varsigma}, v_\kappa \in \mathcal{N}(v_\iota) \},
%	\mathbf{\bar v}_{\iota}^n = \{   \}_{k=1,\dots,N_{i}}
\end{align}
where $\mathcal{N}(v_\iota)$ denotes the neighbor nodes set of the root node $v_\iota$ and $\mathbf{v}_{\kappa,\varsigma}$ is $\varsigma$th element in the feature vector of node $v_\kappa$. We note the edge weight $\mathbf{e}_{\iota,\kappa}$ will be re-computed for each information propagation process. We then perform separate modeling for the root node and the neighbor nodes:
\begin{align}
	\mathbf{\bar v}_\iota = \mathbf{W}_n \mathbf{\bar v}_{\iota}^n + \mathbf{b}_n + \mathbf{W}_r \mathbf{v}_\iota + \mathbf{b}_r,
\end{align}
where $\mathbf{W}_n, \mathbf{W}_r$ are linear transformations to project the root representation and the aggregated neighbors representation into a common space. $\mathbf{b}_n,\mathbf{b}_r$ are the bias terms. This schema further incorporates an element-wise function for re-weighting the importance of each position as well as a short-cut connection:
\begin{align}
	\mathbf{\tilde v}_\iota = \sigma(\mathbf{W}_n^a \mathbf{\bar v}_{\iota}^n + \mathbf{b}_n^a + \mathbf{W}_r^a \mathbf{v}_\iota + \mathbf{b}_r^a) * \mathbf{\bar v}_\iota + \mathbf{v}_\iota.
\end{align}
where $*$ denotes the Hadamard product. Matrices $\mathbf{W}_n^a, \mathbf{W}_r^a$ and the corresponding biases $\mathbf{b}_n^a, \mathbf{b}_r^a$ model the position-wise importance. $\sigma$ denotes the element-wise sigmoid function.

\subsection{Product-aspects Leveraging}

\subsubsection{Knowledge Filtering} Leveraging product aspects as knowledge is an essential part of obtaining the basic product information first and a better understanding of the user-generated product videos later. Different from other kinds of external knowledge (\eg, Knowledge Base and Wikipedia), the aspects of the associated product contain noised values that may hurt the performance of product video understanding. For example, there can be both \texttt{black-white} and \texttt{red-blue} color choices for a certain t-shirt on sell while the buyer/fan may love the \texttt{black-white} one and wear it in the video. We therefore devise a knowledge filtering module based on the hard attention mechanism to filter noised values such as \texttt{red-blue} for each video. Formally, given the nodes features $\{\mathbf{v}_\tau\}_{\tau=1,\dots,N_f * (N_p + 1)}$ in the video graph $G_v$ and product aspect embeddings $\{\mathbf{a}_k\}_{k=1,\dots,N_a}$, we perform knowledge filtering by:
\begin{align}
	& \mathbf{v}^\circ = \frac{1}{N_f * (N_p + 1)} \sum_\tau \mathbf{v}_{\tau}, \\
	&\mathbf{\alpha}_k = \sigma(\mathbf{W}_{h}[ \mathbf{a}_k, \mathbf{v}^\circ] + \mathbf{b}_{h}), \label{eq:importance} \\
	& \bar A = \{ a_{s} | \frac{  \exp(\mathbf{\alpha}_{s}) }{\sum_k \exp( \mathbf{\alpha}_{k}) }  > \beta_{s} \},
\end{align}
where $\bar A$ denotes the filtered aspect set, which includes aspect $a_{s}$ with importance $\mathbf{\alpha}_{s}$ over a certain threshold $\beta_{s}$. We empirically set the threshold to the uniform probability $1 / N_a$. The importance indicator $\mathbf{\alpha}_k$ is computed using a linear transformation $\mathbf{W}_{h}$ and a bias term $\mathbf{b}_{h}$. We use the global (or averaged-pooled) representation $\mathbf{v}^\circ$ of the video graph as the filtering context since we aim to remove aspects that are irrelevant to any part of the video. We add a sigmoid function $\sigma$ to prevent large importance scores, which may lead to a small filtered aspect set after the scores being forward to the $softmax$ function.

\subsubsection{Dynamic Memory Modeling} Previous works that incorporate external knowledge for video description generation \cite{Whitehead_Ji_Bansal_Chang_Voss_2018,Venugopalan_Hendricks_Mooney_Saenko_2016} often leverage the knowledge in the decoding stage, \ie, using pointer mechanism to directly borrow the words/entities from the knowledge document or using attention mechanism to update the decoder hidden state. We propose to progressively retrieve relevant knowledge in the encoding stage, which enables a better understanding of the video for capturing user-preferred product highlights. Specifically, we employ a memory network \cite{Glehre_Chandar_Cho_Bengio_2018,Weston_Chopra_Bordes_2015,Sukhbaatar_Szlam_Weston_Fergus_2015} based approach and enhance it with dynamic memory writing:
\begin{align}
	& \mathbf{\bar a}_{s} = \alpha_{s} * (\mathbf{W}_a \mathbf{a}_{s} + \mathbf{b}_a) + (1 - \alpha_{s}) * (\mathbf{W}_{gl} \mathbf{v}^\circ + \mathbf{b}_{gl} ), \label{eq:memwrite} \\
	& \mathbf{\omega}_{\tau,s} = \mathbf{W}_\omega \tanh(\mathbf{W}_m [\mathbf{v}_\tau, \mathbf{\bar a}_{s}] + \mathbf{b}_m), \\
	& \mathbf{\hat v}_\tau = \gamma * \mathbf{v}_\tau +  \sum_{s} \mathbf{\bar \omega}_{\tau,s} * \mathbf{\bar a}_{s} , \  \textrm{where} \ \mathbf{\bar \omega}_{\tau,s} = \frac{  \exp( \mathbf{\omega}_{\tau,s} ) }{\sum_{o} \exp( \mathbf{\omega}_{\tau,o} ) }.  \label{eq:a3}
\end{align}
where the memory writing process (Equation \ref{eq:memwrite}) borrows the importance factor $\alpha_{s}$ from Equation \ref{eq:importance} (Note that the importance factor $\alpha_{s}$ is in the range $(0, 1)$ after the $\sigma$ function.) This process helps inhibit relatively irrelevant aspect information (with smaller $\alpha_{s}$) and enliven the more relevant ones (with larger $\alpha_{s}$). $\gamma$ controls to what extend the final representation $\mathbf{\hat v}_\tau$ depends on the initial representation $\mathbf{v}_\tau$ and we empirically set it to $0.5$.

\subsection{Progressive Inference Encoder}

Since the spatial-temporal inference module and the knowledge leveraging module updates the node representation without modifying the graph structure, we can easily stack multiple STI modules and multiple KL modules. Poet builds the inference encoder by progressively and alternatively performing STI and KL as depicted in Figure \ref{fig:schema}. In such a design, we aim to not only obtain higher-order graph reasoning (i.e., with access to remote neighbors) but also propagate the leveraged knowledge to the whole graph by the following STI modules. We denote the combination of one STI and one KL as one graph reasoning layer. We use two-layers graph reasoning in the experiment and we observe stacking more layers, which may make node representations over-smoothing and not distinct (i.e., all nodes contain similar information), will lead to a minor performance drop.

%By stacking multiple spatial-temporal inference modules and multiple knowledge leveraging modules, we construct the product-oriented video captioning encoder of \textit{Poet}. By \textit{product-oriented}, we mean the product-part based video graph representation and leveraging product aspects for a better understanding of user-generated videos. [TODO, Add more details Reviewer 2]

\subsection{Decoder}

Following many previous works \cite{Zhu_Jiang_2019,Liu_Ren_Yuan_2018}, we build the decoder with the RNN (here we use gated recurrent unit $GRU$ \cite{Cho_Merrienboer_Bahdanau_Bengio_2014}) and soft attention mechanism. We first initialize the hidden state of $GRU$ as the global representation of the knowledge-aware video graph:
\begin{align}
	\mathbf{h}_0 = \mathbf{v}^\circ = \frac{1}{N_f * (N_p + 1)} \sum_\tau \mathbf{v}_{\tau},
\end{align}
For each decoding step $t$, we attend to each node inside the video graph and aggregate the visual cues using the weighted sum:
\begin{align}
	& \mathbf{\varrho}_{\tau} = \mathbf{W}_\varrho \tanh(\mathbf{W}_{md} [\mathbf{v}_\tau, \mathbf{h}_t] + \mathbf{b}_{md}), \\
	& \mathbf{\hat h}_t =   \sum_{\tau} \mathbf{\bar \varrho}_{\tau} * \mathbf{v}_{\tau} , \mathrm{where} \  \mathbf{\bar \varrho}_{\tau} = \frac{  \exp( \mathbf{\varrho}_{\tau} ) }{\sum_{\tau} \exp( \mathbf{\varrho}_{\tau} ) },
\end{align}
where $ \mathbf{W}_\varrho, \mathbf{W}_{md}$ are linear transformations and they together model the additive attention \cite{Cho_Merrienboer_Bahdanau_Bengio_2014} with the bias term $\mathbf{b}_{md}$. $\mathbf{\bar \varrho}$ is the attention weights. The $t$th decoding process can be formulated as:
\begin{align}
	\mathbf{h}_{t+1} = GRU(\mathbf{h}_{t}, [\mathbf{\hat w}_t, \mathbf{\hat h}_t]),
\end{align}
where $\mathbf{\hat w}_t$ denotes the embedding of the predicted word $w_t$ at step $t$. For training objectives, we take the standard cross-entropy loss:
\begin{align}
	\mathcal{L} = - \sum_{t} \log p\left(w_t\right).
\end{align}

%% file: 5.Experiments.tex
\section{Experiments}

\subsection{Product-oriented Video Datasets} \label{Data Collection}

 % Buyer-generated videos and descriptions are originally from the product comment area, and buyers often focus on the general appearance, salient characteristics, and emotions. Fan-generated videos and descriptions are produced by enthusiasts such as fashionistas,
 
 % Besides product recommendation, e-commerce platforms will also recommend product-related videos that accompany a meaningful description and a link to the relevant product.

% 为什么收集
\vpara{Data Collection} We collect two large-scale product-oriented video datasets, \ie, buyer-generated fashion video dataset (BFVD) and fan-generated fashion video dataset (FFVD) for \textit{product-oriented} video captioning research. Data samples from both datasets are collected from Mobile Taobao. We collect the videos, the descriptions, and the associated product aspects to form the datasets. Each recommended video has been labeled as buyer-generated or fan-generated by the platform. These two kinds of data are originally generated by users with different background knowledge and intentions. Buyers often focus on the general appearance, salient characteristics, and emotions while descriptions generated by fans often reflect deep insights and understandings about the products. Therefore, we regard these two kinds of videos as individual datasets since they may pose different challenges for modeling. Data collected from real-world scenario may contain noises and we select videos with PV (page views) over 100,000 and with CTR (click through rate) larger than 5\%. Videos and descriptions of high quality are more likely to be recommended (more PVs) and clicked (bigger CTR).

%Recommended videos and descriptions have already been filtered by the platform for business profit, and thus BFVD and FFVD are of high quality.

%--------------------------------table-------------------------

\begin{table}[t]
%\begin{strip}
\centering
\caption{
Comparing BFVD and FFVD with exiting video-to-text datasets (e-comm stands for e-commerce).
}
\setlength{\tabcolsep}{2.5pt}
\setlength\doublerulesep{0.5pt}
%\footnotesize  
%\small
\begin{tabular}{l|cccccc}

%\cline{1-9}
% <<<

\multicolumn{1}{c|}{Dataset} &    Domain      &   \#Videos    &   \#Sentence & \#Vocab  &  Dur(hrs)   \\
\hline \hline

MSVD \cite{Chen_Dolan_2011}            & open          &    1,970        &    70,028  & 13,010    & 5.3           \\
%\hline 

%

%YouCook \cite{Das_Xu_Doell_Corso_2013}  & cooking     &  88  &  2,668    & 2,711  &  2.3 \\
%\hline
TACos  \cite{Regneri_Rohrbach_Wetzel_Thater_Schiele_Pinkal_2013}   &  cooking    &  123  &  18,227    & 28,292 &  15.9
        \\
%\hline
TACos M-L \cite{Rohrbach_Rohrbach_Qiu_Friedrich_Pinkal_Schiele_2014}  &  cooking  & 185   &  14,105    & - &  27.1
\\
%\hline
MPII-MD \cite{Rohrbach_Rohrbach_Tandon_Schiele_2015}  &  movie      &  94  &  68,375    & 24,549 &  73.6
\\
%\hline
M-VAD \cite{Torabi_Pal_Larochelle_Courville_2015}  &  movie        &  92  &  55,905     & 18,269  &  84.6
\\
VTW \cite{Zeng_Chen_Niebles_Sun_2016}  &  open        &  18,100    &  -   & 23,059  &  213.2
\\
MSR-VTT \cite{Xu_Mei_Yao_Rui_2016}  &  open  & 7,180    & 200,000     & 29,316 &  41.2
\\
Charades \cite{Sigurdsson_Varol_Wang_Farhadi_Laptev_Gupta_2016}  &  human     &  9,848   & - & -  &  82.01
\\
ActivityNet \cite{Krishna_Hata_Ren_FeiFei_Niebles_2017}  &  activity     &  20,000   & 10,000 & -  &  849
\\
DiDeMo \cite{Hendricks_Wang_Shechtman_Sivic_Darrell_Russell_2017}  &  open     &  10,464   & 40,543 & -  &  -
\\
YouCook2 \cite{Zhou_Xu_Corso_2018}  &  cooking     &  2,000   & - & 2,600  &  175.6
\\
VATEX \cite{Wang_Wu_Chen_Li_Wang_Wang_2019}  &  open     &  41,300   & 826,000 & 82,654  &  -
%\\
%Bilibili  &  live      & 2,361    & 895,929  & 4,860,246 & -  &   113.84
\\
\hline
\hline
BFVD  &  e-comm   & 43,166    & 43,166  & 30,846  &  140.4 \\
FFVD  &  e-comm   & 32,763    & 32,763  & 34,046  &  252.2
\\
\hline
%\\
%\\

\end{tabular}
\vspace{-0.4cm}
\label{table:Data}
\end{table}

%---------------------------table end-------------------------

% Previous datasets mainly focus on cooking, movie, human activity, and open domain for general description generation.

% , which makes the problem even more challenging since models have less chance to easily fit the data samples.

% 统计信息
\vpara{Data Statistics} As a result, we collect 43,166 <video, aspects, description> triplets in BFVD and 32,763 triplets in FFVD. The basic statistics and comparison results with other frequently used video captioning datasets are listed in Table \ref{table:Data}. The distinguishing characteristics of BFVD and FFVD are 1) these datasets can be viewed as an early attempt to promote video-to-text generation for the domain of e-commerce. 2) Concerning the total number and total length of videos, BFVD and FFVD are among the largest. 3) As for language data, BFVD and FFVD contain a large number of unique words, and the ratio of $\frac{\#Vocab}{\#Setences}$ is among the largest. These statistics indicate that BFVD and FFVD contain abundant vocabulary and little repetitive information. 4) BFVD and FFVD associate corresponding product aspects, which permit not only the product-oriented video captioning research as we do but also a broad range of product-related research topics such as multi-label video classification for e-commerce.

% 其他信息 (Preprocessing, Data splits)
% 看是否有空间写完整的preprocessing，如果没有写 Data Splits
\vpara{Data Preprocessing} For descriptions, we remove the stop words and use Jieba Chinese Tokenizer \footnote{\url{https://github.com/fxsjy/jieba}} for tokenization. To filter out those noised expressions such as brand terms and internet slang terms, we remove tokens with frequency less than 30. Descriptions with tokens more than 30 will be shortened, and the max length for product aspects is 12. Following the standard process, we add a \texttt{<sos>} at the beginning of each description and a \texttt{<eos>} at the last.

For videos, we inspect 30 frames for each video and extract the product-part features for each frame using this detector \cite{liu2018deep}\footnote{\url{https://github.com/fdjingyuan/Deep-Fashion-Analysis-ECCV2018}} pre-trained on this dataset \cite{Liu_Yan_Luo_Wang_Tang_2016}. For each frame, the product-part detector will produce a prediction map with the same width/height as the input frame. Each pixel value at a particular part-channel (note that the number of channels is the same as the number of product parts) indicates how likely this pixel is belonging to this part. We then apply the prediction map on the activations obtained from an intermediate layer (by resizing the probability map first, apply softmax on this map to obtain probability weights and finally weighted sum over the intermediate activations per part channel) to obtain part features. Specifically, we use the activations from \textit{pooled\_5} and result in $8 \times 64$ features vectors for 8 product-parts. We mean-pool the activations from layer \textit{conv4} as the representation for each frame.

For train/val/test split, we employ a random sampling strategy and adopt 65\%/5\%/30\% for the training set, validation set, and testing set, respectively. Therefore, we have 21,554/1,658/9,948 data samples for FFVD and 28,058/2,158/12,950 data samples for BFVD.

%--------------------------------table---------------------
\begin{table*}[t]
%\begin{strip}
\centering
\caption{
     Qualitative results of the proposed \textit{Poet} with diverse competitors. Comparisons concern generation quality (NLG metrics), product aspect capturing, and generation diversity. \textit{Poet} achieves the best results on two \textit{product-oriented} video captioning datasets.
}
\setlength{\tabcolsep}{6.5pt}
\setlength\doublerulesep{0.5pt}
%\footnotesize  
%\small
\begin{tabular}{c|l|cccc|c|cc}
%\hline
%\toprule
&  &   \multicolumn{4}{c}{NLG Metrics} & \multicolumn{1}{c}{Aspect} & \multicolumn{2}{c}{Lexical Diversity}  \\
%\cline{1-9}
% <<<
\multicolumn{1}{c|}{Dataset}  & \multicolumn{1}{c|}{Methods}    & BLEU-1        & METEOR   &ROUGE\_L          & CIDEr       &   Prediction  &  $n=4 (\times 10^5)$  &   $n=5 (\times 10^6)$        \\
%\midrule
%

\hline \hline
\multirow{7}{*}{BFVD} & AA-MPLSTM          & 11.31 & 6.02 & 10.08 & 9.76   & 54.31          &   2.94   & 3.20            \\
& AA-Seq2Seq   & 11.96 & 6.14 & 11.05 & 11.67  & 54.85   & 4.74  &  4.52   \\
& AA-HRNE   & 11.82 & 5.98 & 10.23 & 11.86   & 55.98   &  5.02  &  4.73    \\
& AA-SALSTM     & 11.78 & 5.88 & 10.18 & 11.57   & 55.93   &  5.10 & 4.90
        \\
& AA-RecNet   & 11.17 & 6.01 & 11.05 & 11.67  & 54.94   &  5.06  &  4.92
        \\
& Unified-Transformer   & 11.28 & 6.32 & 10.43 & 12.66 & 55.12   &  3.35  &  2.91 \\
& PointerNet & 12.09 & 6.34 & 11.19 & 12.58 &  56.01  &   \textbf{5.36}  & 5.02
        \\ \cmidrule{2-9} \cmidrule{2-9}
& \iffalse\rowcolor{SeaGreen1!20!}\fi \Our  & \textbf{14.55}  & \textbf{7.11}  & \textbf{12.13} & \textbf{13.48} & \textbf{56.69} & 5.16 &  \textbf{5.10}\\

\hline
\multirow{7}{*}{FFVD} & AA-MPLSTM          & 14.52 & 7.96 & 13.85 & 17.38   &     61.63       &  3.15   &   3.22            \\
& AA-Seq2Seq   & 14.77 & 7.87 & 13.74 & 18.54  &  62.01   &   4.08  &  3.69 \\
& AA-HRNE   & 13.58 & 6.75 & 12.06 & 20.10   &   60.39   &  4.32   &  3.88  \\
& AA-SALSTM    & \textbf{16.25} & 7.72 & 14.63 & 19.46   &  62.17   &  4.58  &  4.20
        \\
& AA-RecNet    & 15.11 & 8.03 & 14.18 & 19.08  &   62.21   & 4.45  &  4.02
        \\
& Unified-Transformer   & 14.39 & 7.42 & 13.45 & 21.00 &  62.01   & 3.39  & 2.90
        \\
& PointerNet   & 15.28 & 7.77 & 14.02 & 18.85 & 61.30  & 4.40 &  3.99
        \\ \cmidrule{2-9} \cmidrule{2-9}
& \iffalse\rowcolor{SeaGreen1!20!}\fi \Our  & 16.04  & \textbf{8.06}  & \textbf{14.82} & \textbf{21.71} & \textbf{62.70}  & \textbf{4.60}  & \textbf{4.25}  \\
\hline
% >>> 0.1604	0.0806	0.1482	0.2181

\end{tabular}
%\medskip
\label{tab:qualitative}
\vspace{-0.3cm}
\end{table*}
%
%--------------------------------table end---------------------
%

\subsection{Evaluation Measurements}

% We are analyzing our system in terms of frequently used natural language generation metrics and the evaluation framework used in product description generation \cite{Chen_Lin_Zhang_Yang_Zhou_Tang_2019}, including product aspects prediction and lexical diversity evaluation.

% Reference-based measurements directly compare the generated sentences with the ground-truth sentences (as reference).

\begin{sloppypar}
	\vpara{Natural Language Generation Metrics} Concretely, we adopt four numerical assessment approaches, BLEU-1 \cite{papineni2002bleu} for sanity check, METEOR \cite{banerjee2005meteor} based on unigram precision and recall, ROUGE\_L \cite{lin2004rouge} based on the longest common subsequence co-occurrence, CIDEr \cite{vedantam2015cider} based on human-like consensus. With the user-written sentences as references, these measurements can help evaluate the generation fluency as well as whether the generation model captures the user-preferred highlights depicted in the video.
\end{sloppypar}

\vpara{Product Aspects Prediction} Since leverging product aspects as knowledge is essential for product video understanding, it is necessary to evaluate how well the generation model captures such information. We follow the evaluation protocol proposed in KOBE \cite{Chen_Lin_Zhang_Yang_Zhou_Tang_2019} and view the aspects prediction accuracy as the indicator.

\vpara{Lexical Diversity} The above measurements mainly concern the generation quality (fluency, highlight capturing, and aspects capturing), and they cannot explicitly evaluate the generation diversity. As we know, repetitive phrases and general descriptions can be less attractive for potential buyers. Therefore, as in KOBE \cite{Chen_Lin_Zhang_Yang_Zhou_Tang_2019}, we view the number of unique n-grams of the sentences generated when testing as the indicator of the generation diversity. We empirically choose 4-grams and 5-grams, following KOBE.

\subsection{Comparison Baselines}

To evaluate the effectiveness of \textit{Poet}, we compare it with various video description baselines. Since most baselines concern only video information, we re-implement these baselines and add separate encoders (with a similar structure to their video encoders) for product aspect modeling.

\begin{enumerate}
	\item \textit{AA-MPLSTM}. Aspect-aware MPLSTM (originally \cite{Venugopalan_Xu_Donahue_Rohrbach_Mooney_Saenko_2015}) employs two mean-pooling encoders and concatenate the encoded vectors as the decoder input.
	\item \textit{AA-Seq2Seq}. Aspect-aware Seq2Seq (originally \cite{Venugopalan_Rohrbach_Donahue_Mooney_Darrell_Saenko_2015}) uses an additional RNN encoder for aspect modeling.
	\item \textit{AA-SALSTM}. Aspect-aware SALSTM (originally \cite{Yao_Torabi_Cho_Ballas_Pal_Larochelle_Courville_2015}) equip the RNN decoder in AA-S2VT with soft attention.
	\item \textit{AA-HRNE}. Aspect-aware HRNE (originally \cite{Pan_Xu_Yang_Wu_Zhuang_2016}) differs the AA-S2VT by employing hiearachical encoders.
	\item \textit{AA-RecNet}. For Aspect-aware RecNet, we re-construct not only the frame features (as in RecNet \cite{Wang_Ma_Zhang_Liu_2018}) but also the aspect features. Also, there is an additional aspect encoder.
	\item \textit{Unified-Transformer}. We modify the Unified Transformer \cite{Ma_Cui_Dai_Wei_Sun_2019} by replacing the live comments encoder with an aspect encoder.
	\item \textit{PointerNet}. We equip the Seq2Seq model with the entity pointer network, proposed by Whitehead \etal \cite{Whitehead_Ji_Bansal_Chang_Voss_2018}, for product aspect modeling.
\end{enumerate}

\subsection{Performance Analysis}

In this section, we examine the empirical performance of \textit{Poet} on two product-oriented video datasets, \ie, BFVD and FFVD. We concern a couple of the following perspectives.

\vpara{Overall generation quality}. By generation quality, we mean both the generation fluency and whether the generated sentences capture user-preferred product characteristics. Referenced-based natural language metrics can help reflect the performance since they directly compare the generated sentences to what the video uploaders describe. In a nutshell, the clear improvement over various competitors and across four different metrics demonstrate the superiority of the proposed \textit{Poet} (as shown in Table \ref{tab:qualitative}, NLG metrics). Specifically, on BFVD, we obtain +2.46 BLEU (relatively 20.3\%) and +0.77 METEOR (relatively 12.1\%) improvement over the best (PointerNet) among the competitors. For FFVD, we observe that fan-generated videos sometimes concern collections of clothes and present them for a specific theme (such as \texttt{dressing guide} or \texttt{clothes that make you look fit}) while only one of the clothes (in the same video) is associated with product aspects. Such a phenomenon may introduce noise and hurt the performance of models designed for single-product modeling. Nevertheless, \textit{Poet} achieves the best performance across the most metrics. We attribute the clear advantage to \textbf{1) product-oriented video graph modeling.} \textit{Poet} represents product parts across frames as spatial-temporal graphs, which can better capture the dynamic change of these characteristics along the timeline and find out those distinguishing highlights that are preferred by the user (\eg, distinguishing characteristics of one product part can be highlighted in higher frequencies or with close-up views). In contrast, previous models (including the RNN-based and the transformer-based) are inferior in capturing such characteristics since they either model only the frame-level features without fine-grained analysis or only model the videos in a sequential way. \textbf{2) Knowledge enhanced video analysis.} \textit{Poet} firstly perform hard attention to remove \textit{noise aspects} that are of no use for video analysis and then perform dynamic memory writing/attending to progressively enhance the spatial-temporal inference process. This design can be superior over other designs like concatenation or the complex PointerNet design. We will further analyze different knowledge incorporation methods in \textit{Aspect Capturing} and \textit{Ablation studies}.

\vpara{Aspect Capturing} Knowledge incorporation methods can be categorized into three sub-groups: 1) \textbf{AA- methods.} We transform a video captioning method into a AA-method by adding a separate encoder, which has a similar structure to the existing visual encoder (such as the hierarchical RNN encoder in HRNE), for the product aspect modeling. We concatenate the encoded aspects feature and the encoded visual feature as the initial decoder input. 2) \textbf{Decoding-oriented methods.} Unified-transformer and PointerNet are decoding-oriented knowledge incorporation methods, which introduce the external knowledge in the decoding stage as an implicit or explicit reference. Unified-transformer implicitly utilizes the knowledge using the multi-head attention mechanism before prediction. PointerNet explicitly view the attended entities (product aspects) as candidate words for prediction and combine the attention weights and the vocabulary probability distribution before prediction. 3) \textbf{Poet}: the analysis-oriented method which aims to obtain a better understanding of the videos with the external knowledge as guidance. The aspect capturing scores are shown in Table \ref{tab:qualitative} (Aspect Prediction). It can be seen that 1) the analysis based method (\textit{Poet}) achieves the best performance on two datasets. As we illustrated earlier, the \textit{product-oriented} video captioning requires a fine-grained analysis of distinguishing characteristics depicted in the video, and the product aspects can better serve as the prior background knowledge to obtain such kind of analysis. 2) The decoding-oriented methods cannot beat the AA- methods on FFVD. This is a reasonable result since the product aspect can be associated with only one cloth in the fan-generated video (as we stated in the \textit{Overall Generation Quality}) and thus directly \textit{borrowing} the aspect words in the decoding may introduce unnecessary local biases.

%--------------------------------table-------------------------

\begin{table}[!t]
\centering
\caption{
Human judgements on the proposed \textit{Poet} and two typical architectures concerning three task-oriented indicators.
}
\setlength{\tabcolsep}{5.5pt}
\setlength\doublerulesep{0.5pt}
\begin{tabular}{l|ccc}

%\cline{1-9}
% <<<

\multicolumn{1}{c|}{Models} &    Fluency      &   Diversity    &    Overall Quality  \\
\hline \hline

%\hline
AA-RecNet   &  2.73  & 3.49    & 3.04
\\
%\hline
AA-Transformer   &  2.66  & 3.37    & 2.95
\\
\hline
\hline
\Our   & \textbf{2.88}  & \textbf{3.59}    & \textbf{3.15} 
\\

\hline

\end{tabular}
%\vspace{-0.6cm}
\label{table:HumanEval}
\end{table}

%---------------------------table end-------------------------

\vpara{Generation Diversity} The number of generated unique n-grams of different methods are listed in Table \ref{tab:qualitative} (Lexical Diversity). Besides the generation quality, \textit{Poet} can also generate relatively diversified sentences. The PointerNet achieves the best concerning 4-grams in BFVD, which is reasonable since they externally consider the aspects as candidate decoding words.

%--------------------------------fig---------------------
\begin{figure*}[t] \begin{center}
    \includegraphics[width=\textwidth]{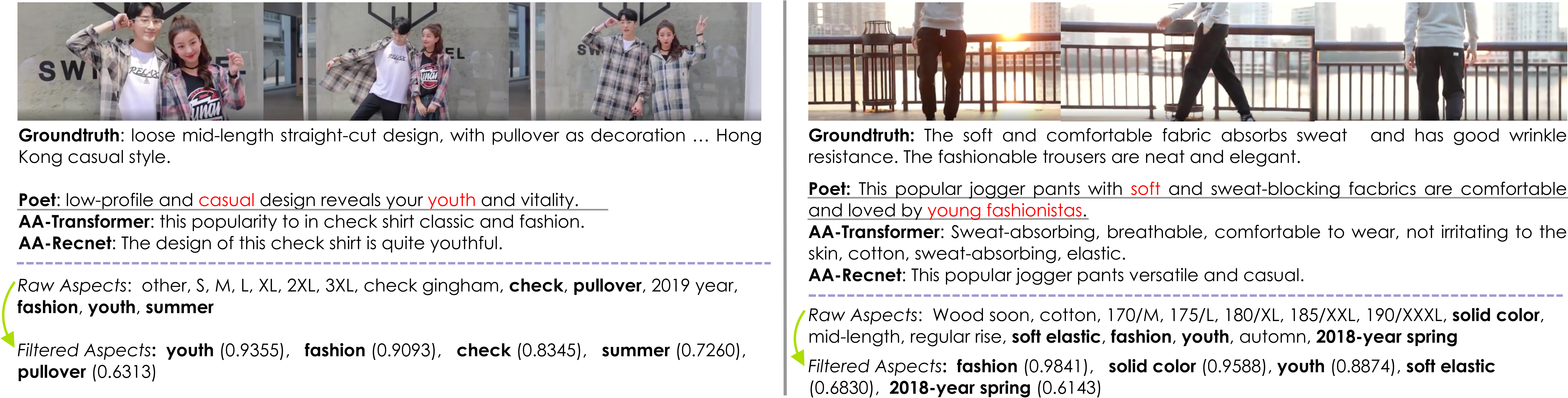}
    \caption{
    	Generation samples of \textit{Poet}, AA-Transformer and AA-Recnet on the FFVD testing set. We present the filtered aspects and the corresponding scores in the proposed knowledge leveraging module.
	}
\vspace{-0.4cm}
\label{fig:case}
\end{center} \end{figure*}
%--------------------------------fig end---------------------

\vpara{Human Evaluation} We agree with AREL \cite{Wang_Chen_Wang_Wang_2018} that human judgement is essential for stable evaluation especially when the captions are highly diverse and creative. Following the human evaluation protocol of KOBE \cite{Chen_Lin_Zhang_Yang_Zhou_Tang_2019} and Li \etal \cite{Li_Monroe_Shi_Jean_Ritter_Jurafsky_2017}, we randomly select 1,000 instances from the testing set and distribute them to human annotators. The results are listed in Table \ref{table:HumanEval}. Compared to the typical Transformer-based design and RNN-based design (AA-RecNet), \textit{Poet} generates more fluent (fluency +0.22/+0.15) and diversified (diversity +0.22/+0.10) descriptions. The indicator \textit{Overall Quality} reflects whether the descriptions capture the product characteristics the video uploader hightlights in the video. In terms of this metric, \textit{Poet} still demonstrates a clear performance improvement over AA-Transformer/AA-RecNet by +0.20/+0.11.

%--------------------------------table---------------------
\begin{table}[!t]
%\begin{strip}
\centering
\caption{
     Ablation study on the generation quality of Knowledge Leveraging module and the pointer mechanism.
}
\setlength{\tabcolsep}{3.5pt}
\setlength\doublerulesep{0.5pt}
%\footnotesize  
%\small
\begin{tabular}{c|l|cccc}
%\hline
%\toprule
%&  &   \multicolumn{4}{c}{NLG Metrics}  \\
%\cline{1-9}
% <<<
\multicolumn{1}{c|}{Dataset}  & \multicolumn{1}{c|}{Methods}    & BLEU-1        & METEOR   &ROUGE\_L     & CIDEr       \\
%\midrule
%

\hline \hline
\multirow{3}{*}{BFVD} & \iffalse\rowcolor{SeaGreen1!20!}\fi \Our  & \textbf{14.55}  & \textbf{7.11}  & \textbf{12.13} & \textbf{13.48} \\
& \ \ + pointer  & 13.26  & 6.60  & 11.53 & 13.18  \\
& \ \ \ \ \ \  - KL  & 12.43  & 6.48  & 10.86 & 12.25 \\

\hline
\multirow{3}{*}{FFVD} & \iffalse\rowcolor{SeaGreen1!20!}\fi \Our  & 16.04  & \textbf{8.06}  & \textbf{14.82} & \textbf{21.71}  \\
& \ \ + pointer  & \textbf{16.13}  & 7.79  & 14.50 & 20.57  \\
& \ \ \ \ \ \  - KL & 15.53  & 7.89  & 14.18 & 19.73   \\
\hline
% >>> 0.1604	0.0806	0.1482	0.2181

\end{tabular}
%\medskip
\label{tab:ablation_knowledge}
\vspace{-0.3cm}
\end{table}
%
%--------------------------------table end---------------------
%

\vpara{Case Study} Figure \ref{fig:case} shows two generation cases on the FFVD testing set. In summary, \textit{Poet} generates more fluent and complete sentences than the AA-Transformer and AA-Recnet, which are typical architectures of transformer-based and RNN-based models, respectively. For example, the phrases (such as "\texttt{to in}") generated by AA-Transformer in the first case are confusing, and the whole sentence is incomplete. Besides the generation fluency, \textit{Poet} can generate sentences that better capture the product aspects. In the second case, the phrases "\texttt{soft}" and "\texttt{young fashionistas}" are derived from the aspects "\texttt{soft elastic}", "\texttt{youth}", and "\texttt{fashion}". To further demonstrate the effectiveness of the proposed knowledge leveraging module, we extract and present the filtered aspects with the corresponding scores. We observe that the proposed KL module successfully filter those aspects of no use, such as product sizes (\texttt{XL}) and the release year \textit{2019} in the first case, and the scores of remaining aspects are consistent to the contribution to the final description. Also, Poet can generate creative while accurate words (\eg, \texttt{casual}) beyond the input aspects set based on its understanding of the video.

%some title generation samples from Gavotte and the other two competitors, \ie, M-RecNet, and M-Livebot. Similar to the results in human evaluation, Gavotte can generate more fluent and attractive titles. Specifically, while the title of M-Recnet in the first case is less informative and the title of M-LiveBot in the second case is unfinished or broken, Gavotte generates smooth and meaningful title with the popular buzzwords -- "\texttt{steal the show}". The results further demonstrate that Gavotte can recognize granular-level details like "\texttt{ripped}", clothes-level design like "\texttt{jeans}", frame-level product-background interaction effect like "\texttt{steal the show}", and video-level story-line topic like "\texttt{dresses this way}". 

\subsection{Ablation Studies}

We conduct ablation studies to verify the effectiveness of proposed modules within \textit{Poet}. We mainly concern the following two issues:

\vpara{When modeling the video as a graph, does the knowledge leveraging module outperform the pointer mechanism?} To answer this question, we construct two models, \ie, the Poet+pointer model, which directly adds the pointer mechanism to \textit{Poet}, and the Poet+pointer-KL model, which remove the proposed knowledge leveraging module from Poet+pointer model. The experiment results on two datasets are listed in Table \ref{tab:ablation_knowledge}. It can be seen that 1) adding pointer mechanism may hurt the performance in most cases. This further demonstrates the superiority of the analysis-oriented knowledge incorporation method for \textit{product-oriented} video captioning. 2) removing the knowledge leveraging module leads to a clear performance drop (-0.93 CIDEr and relatively 7\% in BFVD), which shows the effectiveness of the proposed module.

\vpara{Do all the proposed modules improve the generation quality?} we surgically remove the proposed modules and individually test the performance on BFVD. Table \ref{table:ablation_module} shows the numeric results. We note that removing the KL (knowledge leveraging) module means ignoring the external product aspects totally. The result indicates the merit of leveraging the KL module to enhance the fine-grained video analysis. By "-SPI", we replace the SPI (spatial-temporal inference) module by the popular Graph Convolutional Networks \cite{Kipf_Welling_2017}. The improvement over the GCN verifies the effectiveness of the spatial-temporal inference module.

%--------------------------------table-------------------------

\begin{table}[!t]
\centering
\caption{
Ablation study of the proposed \textit{Poet} by surgically removing controlling components.
}
\setlength{\tabcolsep}{2.5pt}
\setlength\doublerulesep{0.5pt}
\begin{tabular}{l|cccc}

%\cline{1-9}
% <<<

  \multicolumn{1}{c|}{Models}    & BLEU-1       & METEOR   &ROUGE\_L          & CIDEr       \\
\hline \hline

 \Our      & \textbf{14.55}  & \textbf{7.11}  & \textbf{12.13} & \textbf{13.48}             \\

\hline 
 \ \  - KL  &  11.85  &   6.34  & 10.58  & 12.09
\\
%\hline
  \ \ \ \  - STI   &  11.90  & 6.09    & 10.31  & 9.52 
\\
%\hline
\hline
%\multirow{2}{*}{Input}  & \ \  - Video            & 2.10          &    13.18        &    20.54 & 50.99             \\
%%\hline 
%   & \ \ \ \  - Attribute  & 0.81   & 9.50    &   14.36 & 29.97   \\
%\hline

\end{tabular}
\vspace{-0.4cm}
\label{table:ablation_module}
\end{table}

%--------------------------------table end-------------------------

%% file: 6.Conclusion.tex
\section{Conclusion}

In this paper, we propose to narrate the user-preferred product characteristics depicted in user-generated product videos, in natural language. Automating the video description generation process helps video recommendation systems in e-commerce to leverage the massive user-generated videos for product promotion. We propose a novel framework named \textit{Poet} to perform knowledge-enhanced spatial-temporal inference on product-oriented video graphs. We conduct extensive experiments including qualitative analysis, ablation studies, and numerical measurements concerning generation quality/diversity. Experiment results show the merit of video graph modeling, the proposed spatial-temporal inference module, and the knowledge leveraging module for the \textit{product-oriented} video captioning problem. We collect two user-generated fashion video datasets associated with product aspects to promote not only the \textit{product-oriented} video captioning research, but also various product-oriented research topics such as product video tagging.

\section{ACKNOWLEDGMENTS}

\begin{sloppypar}
The work is supported by the NSFC (61625107, 61751209, 61836002), National Key R\&D Program of China (No. 2018AAA0101900, No. 2018AAA0100603), Zhejiang Natural Science Foundation (LR19F020006), Fundamental Research Funds for the Central Universities (2020QNA5024),  and a research fund supported by Alibaba.
\end{sloppypar}